\pdfoutput=1
\documentclass[11pt]{article}

\usepackage[final]{acl}

\usepackage{times}
\usepackage{latexsym}

\usepackage[T1]{fontenc}
\usepackage[utf8]{inputenc}

\usepackage{microtype}

\usepackage{inconsolata}

\usepackage{graphicx}
\usepackage{multicol}
\usepackage{multirow}
\usepackage[ruled,vlined,linesnumbered]{algorithm2e}
\usepackage{xcolor}         % colors
\usepackage{hyperref}       % hyperlinks
\usepackage{url}            % simple URL typesetting
\usepackage{booktabs}       % professional-quality tables
\usepackage{amssymb}
\usepackage{amsmath}
\usepackage{adjustbox}
\usepackage{ulem}

\definecolor{green1}{rgb}{0,0.5,0}
\usepackage[most]{tcolorbox}
\usepackage{listings}
\usepackage{xcolor}
\definecolor{softblue}{RGB}{100, 149, 237}
\definecolor{darkpurple}{RGB}{160, 0, 160}
\definecolor{darkred}{RGB}{160, 0, 0}
\definecolor{darkblue}{RGB}{0, 0, 160}
\tcbuselibrary{breakable}
\usepackage[table]{xcolor}
\usepackage{pifont}
\usepackage{threeparttable}
\def\xiao{} % \textcolor{orange}

\def\eg{\emph{e.g.}} 
\def\ie{\emph{i.e.}}

\SetAlCapSkip{0.1em}           % caption 和算法正文之间距离
\SetAlgoNlRelativeSize{-1}     % 行号小一号
\SetAlgoSkip{smallskip}        % 行间距紧凑

\newcommand{\cmark}{\textcolor{green}{\ding{51}}} % 勾选符号
\newcommand{\xmark}{\textcolor{red}{\ding{55}}}   % 叉号

\title{A Knowledge-driven Adaptive Collaboration of LLMs for \\Enhancing Medical Decision-making}

\author{Xiao Wu$^{1,2}$ \quad Ting-Zhu Huang$^{2*}$ \quad Liang-Jian Deng 
  $^2$ \quad Yanyuan Qiao$^3$ \\ {\bf Imran Razzak}$^1$ \quad {\bf Yutong Xie}$^1\thanks{Corresponding author.}$ \\
  $^1$\normalsize{Mohamed bin Zayed University of Artificial Intelligence (MBZUAI)} \\
  $^2$\normalsize{University of Electronic Science and Technology of China} \\
  $^3$\normalsize{Swiss Federal Institute of Technology Lausanne (EPFL)} \\
  \small\tt\{xiao.wu, imran.razzak, yutong.xie\}@mbzuai.ac.ae \quad
  \small\tt tingzhuhuang@126.com \\
  \small\tt liangjian.deng@uestc.edu.cn \quad
  \small\tt yanyuan.qiao@epfl.ch \\
  }

\begin{document}
\maketitle

\begin{abstract}
Medical decision-making often involves integrating knowledge from multiple clinical specialties, typically achieved through multidisciplinary teams. Inspired by this collaborative process, recent work has leveraged large language models (LLMs) in multi-agent collaboration frameworks to emulate expert teamwork. While these approaches improve reasoning through agent interaction, they are limited by static, pre-assigned roles, which hinder adaptability and dynamic knowledge integration.
To address these limitations, we propose KAMAC, a \textbf{Knowledge-driven Adaptive Multi-Agent Collaboration} framework that enables LLM agents to dynamically form and expand expert teams based on the evolving diagnostic context. KAMAC begins with one or more expert agents and then conducts a knowledge-driven discussion to identify and fill knowledge gaps by recruiting additional specialists as needed. This supports flexible, scalable collaboration in complex clinical scenarios, with decisions finalized through reviewing updated agent comments.
Experiments on two real-world medical benchmarks demonstrate that KAMAC significantly outperforms both single-agent and advanced multi-agent methods, particularly in complex clinical scenarios (\textit{i.e.}, cancer prognosis) requiring dynamic, cross-specialty expertise. Our code is publicly available at: \url{https://github.com/XiaoXiao-Woo/KAMAC}.

\end{abstract}

\section{Introduction}
In healthcare, diagnosis, prognosis, and a variety of clinical treatments are guided by medical decision-making processes that require the application of complex medical knowledge~\cite{sutton2020overview}. An individual professional medical perspective is not enough to meet the needs of patients. Multidisciplinary teams (MDTs) or integrated care teams may participate in disease treatment in practical clinical processes~\cite{kodner2002integrated}. 

% Recently, large language models (LLMs) have shown the potential to play the role of clinicians in these medical teams and effectively support medical decision-making~\cite{mdagents, mdteamgpt}.
% % 
% Multi-agent collaboration (MAC) methods based on LLMs play a key role in medical decision-making, enhancing the reasoning ability of LLMs. 
Recently, large language models (LLMs), owing to their powerful reasoning and knowledge synthesis capabilities, have demonstrated promising potential in emulating the roles of clinicians and supporting medical decision-making~\cite{medagents, mdagents, mdteamgpt}. Multi-agent collaboration (MAC) based on LLMs has emerged as a key paradigm, enhancing the reasoning performance of individual agents through collective deliberation.
For instance, \cite{medagents} verified that a training-free collaboration framework in which multiple LLM-based agents simulate a multidisciplinary medical team through role-playing and multi-round discussions, and achieved strong performance across medical question answering (QA) datasets. In addition, \cite{mdteamgpt} further leveraged medical multi-agents and implemented cumulative consultation strategies using retrieval augmentation generation (RAG),
which enhances model outputs by retrieving external medical knowledge to support clinical reasoning and improve diagnostic accuracy.

Some multi-LLM debate frameworks are also closely related to collaboration~\cite{voting, reconcile, cooperation, encouraging}. Among them, a framework for iterative collaboration between agents to make decisions is proposed, which stimulates higher quality answers (compared to a single model) by involving multiple models in the discussion. These works explore the potential application of LLMs and the possibility of their use in medical MDT decisions.

\begin{figure*}
    \centering
    \includegraphics[width=1.0\linewidth]{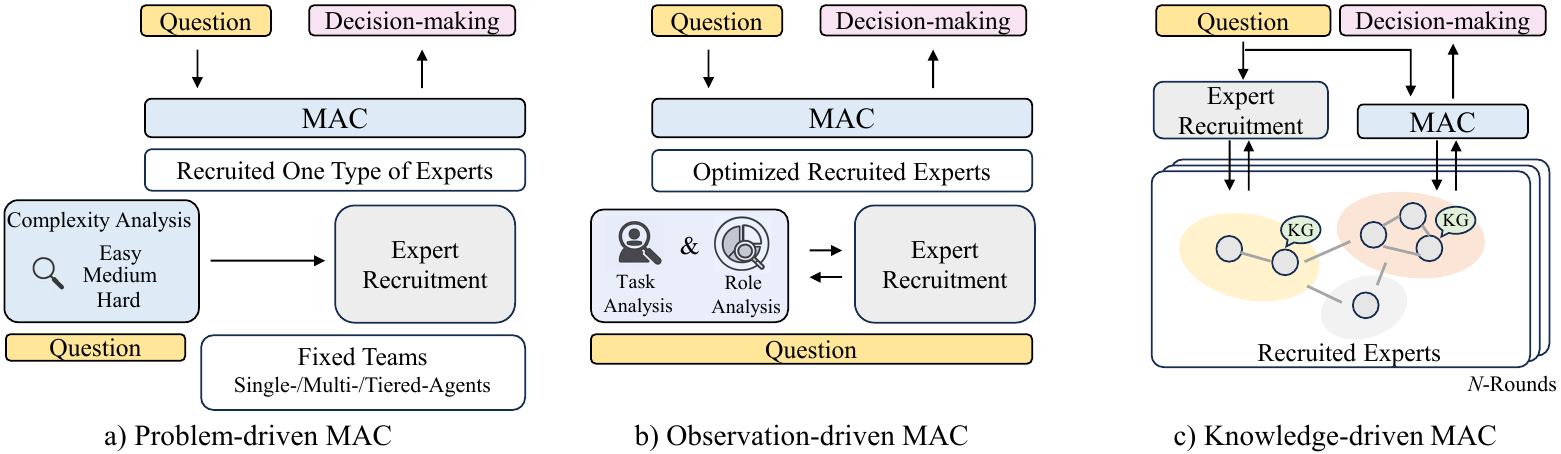}
    \caption{
    \textbf{Comparison of multi-agent collaboration (MAC) strategies in medical decision-making.}
    (a) \textit{Problem-driven MAC}~\cite{mdagents,critique} uses predefined question-complexity tiers (easy, medium, hard) to assemble static single- or multi-tier expert teams
    (b) \textit{Observation-driven MAC}~\cite{agentverse, autoagents} dynamically analyzes task and role characteristics from initial observations to optimize expert recruitment for each question.
    (c) Our proposed \textit{Knowledge-driven MAC} adaptively expands the expert team during discussion by detecting knowledge gaps ({KG}), enabling scalable and flexible collaboration for complex, cross-domain clinical scenarios.}
    \label{fig:comparision}
\end{figure*}

Although these MAC methods enable agents to tackle problems that are difficult or unsolvable by a single agent by learning new contexts and actions through interactions with peers or known information, the challenge remains unresolved. It mainly stems from the use of static, pre-assigned roles based on inherent domain knowledge, which limits the system’s adaptability during collaboration. As discussions progress, each agent tends to produce increasingly fine-grained analyses within its fixed specialty. For example, in the evaluation of a patient presenting with chest pain, a radiology agent may focus solely on imaging findings suggestive of pulmonary embolism, while a cardiology agent may emphasize electrocardiogram (ECG) changes indicative of myocardial infarction. Without a mechanism to reconcile or adapt these perspectives, the collaboration degenerates into a juxtaposition of isolated preferences rather than a convergent diagnostic consensus. This fragmentation undermines the effectiveness of consensus strategies and restricts the system’s ability to dynamically incorporate broader context or cross-domain reasoning.

Recent studies have attempted to improve MAC flexibility by incorporating novel expert recruitment strategies. For instance, problem-driven MAC~\cite{mdagents} (Figure~\ref{fig:comparision}a) assigns expert teams based on question complexity, while observation-driven MAC~\cite{autoagents,agentverse} (Figure~\ref{fig:comparision}b) selects experts according to task and role analysis. However, these methods still rely on static or pre-optimized expert pools and cannot adapt during multi-round interactions. As a result, even when new, fine-grained insights emerge over multiple discussion rounds, no new experts are brought in. The limitations in these works still hinder truly scenario-specific collaboration, especially in dynamic and diverse clinical environments.

To alleviate this, as illustrated in Figure~\ref{fig:comparision}c, 
we propose a \textbf{Knowledge-driven Adaptive Multi-Agent Collaboration (KAMAC)} framework for enhancing medical decision-making. Specifically, KAMAC dynamically increases the number of medical expert team members required for patients by exploring additional expert knowledge during the discussion process. 
KAMAC begins with an initial consultation involving one or more experts. It then engages in a knowledge-driven collaborative discussion, which assesses whether additional expertise is needed by detecting knowledge gaps (KG) and dynamically recruits appropriate experts to fill the knowledge gaps, enabling scalable and flexible collaboration for complex, cross-domain clinical scenarios. Finally, a moderator is responsible for reviewing updated agent comments to complete the decision-making process. Such progressive collaboration and flexible team expansion allow the model to adaptively allocate resources and produce more accurate, context-aware decisions. In contrast to prior methods, the proposed method enables the system to adapt to the evolving clinical treatment in the real world and provide more nuanced and comprehensive support to patients.

Our contributions include three folds:
\begin{enumerate}
    \item We propose the KAMAC framework that dynamically extends a single expert agent into multiple expert agents to form a multi-disciplinary team for medical decision-making.
    \item We design a knowledge-driven collaborative discussion mechanism that enables agents to dynamically expand the team to fill knowledge gaps, aiming to improve adaptability and decision accuracy in complex clinical scenarios.
    \item Extensive experiments on two medical benchmarks, MedQA and Progn-VQA, demonstrating that our KAMAC improves single-agent and advanced multi-agent collaboration frameworks.
\end{enumerate}

\begin{figure*}[ht!]
    \centering
    \includegraphics[width=1.0\linewidth]{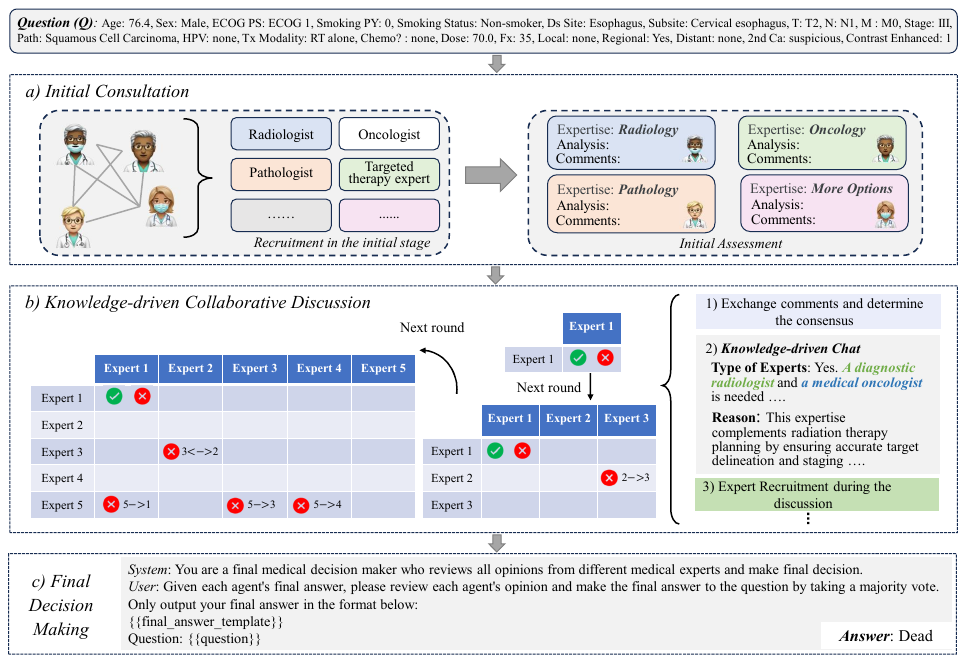}
    \caption{\textbf{Schematic diagram of Knowledge-driven Adaptive Multi-Agent Collaboration (KAMAC) framework for medical decision-making.} 
    The KAMAC includes three parts: (a) Initial Consultation: One or more expert agents (\textit{e.g.}, radiologist, pathologist) are selected based on the clinical question to provide initial assessments; (b) Knowledge-driven Collaborative Discussion: Agents iteratively exchange views to refine reasoning. If a knowledge gap is detected, KAMAC dynamically recruits additional specialists, and the expanded team continues the dialogue until consensus is reached or the round limit is met; and (c) Final Decision Making: A moderator reviews all agent responses and produces the final answer. The symbols \cmark$\,$ and \xmark$\,$ indicate agreement/disagreement with the current expert's comment, respectively. Only when a disagreement occurs, (\xmark{} $i$ -> $j$) or (\xmark{} $i$ <-> $j$) is used to denote a one-way or two-way discussion between expert $i$ and expert $j$, respectively.
    }
    \label{fig:main_fig}
\end{figure*}

\section{Related Work}
Advanced LLMs such as GPT-4~\cite{gpt}, DeepSeek~\cite{deepseek-v3,deepseek-r1}, and Gemini~\cite{gemini} have demonstrated strong reasoning capabilities and have been used as agents with considerable computational investment in various medical tasks such as question answering~\cite{mdagents, medagents}, diagnosis~\cite{generalist}, and report generation~\cite{xraygpt, maira}. We list two main related areas of work:
\paragraph{LLM-Based Agentic Medical Decision-Making} Medical decision-making systems leverage multiple LLM experts, each assigned a predefined clinical specialty to mimic real‐world multidisciplinary teams. Early work demonstrated that consensus among expert agents yields higher diagnostic accuracy than any single model or simple voting schemes~\cite{medagents, encouraging, reconcile}. Some recent works mainly focused on diagnostic findings~\cite{mdagents, mediq} and knowledge integration~\cite{medrag, gpt-medical, mdagents}. MediQ~\cite{mediq} designs a system to seek methods to guide the deepening of interactions between patients and experts. For instance, \cite{mdagents} verified that expert collaboration has better accuracy for medical decision-making than a single expert, showcasing that consensus is superior to a voting strategy in various clinical applications. More recently, MDteamGPT~\cite{mdteamgpt} adds a leader agent, historical dialogues, and RAG to integrate information and supplementation strategies to assist in decision-making.
\paragraph{Multi-Agent Collaboration in Medical Decision-Making} Researchers have demonstrated that multi-agent collaborative research can enhance the reasoning capabilities of these LLMs~\cite{clinicalagent,moa, AgentForest}.  Well-designed strategies can enhance autonomous multi-agent systems for task-solving capabilities, such as debate~\cite{chateval, cooperation}, consensus~\cite{voting, reconcile}, conflict-solving, generation/evolution~\cite{yuan2024evoagent,agentverse, autoagents}, and encouragement~\cite{encouraging, mc_survey}. Some multi-LLM debate frameworks are also closely related to collaboration~\cite{voting, reconcile, cooperation, encouraging}. Among them, a framework for iterative collaboration between agents to make decisions is proposed, which stimulates higher quality answers (compared to a single model) by involving multiple models in the discussion. These works explore the potential application of LLMs and the possibility of their use in medical MDT diagnostics.
Although assigning experts can effectively improve the performance of specific tasks, the rationality of expert assignment in multi-agent collaboration is still insufficient. \cite{autoagents,agentverse} introduces optimal expert generation strategies in the initial expert recruitment stage, but it does not consider the relationship between expert knowledge and cooperation between experts. \cite{yuan2024evoagent} introduces a dynamic evolution strategy for the existing experts but relies on a large initial population and requires additional investment. These limitations make it unsuitable for medical decision-making.

% % Define custom comment styles
% \newcommand\myCommentStyle[1]{\textcolor{green!50!black}{#1}}  % Green style
% % \newcommand\blueCommentStyle[1]{\textcolor{blue}{#1}}          % Blue style
% % Set comment styles for algorithm2e
% \SetKwComment{Comment}{$\triangleright$ }{\myCommentStyle{#1}}  % Green comments
% % \SetKwComment{Atitle}{$\star$ }{\blueCommentStyle{#1}}          % Blue comments

% \normalem

\SetCommentSty{myCommentStyle}
\newcommand\myCommentStyle[1]{\textcolor{green1}{#1}}
\SetKwComment{Comment}{$\triangleright$ }{}
\normalem

\begin{algorithm*}[ht!]
        \caption{Knowledge-driven Adaptive Multi-Agent Collaboration (KAMAC) Decision-making}
        \label{alg:kac}
        \KwIn{Problem $\mathcal{Q}$}
        \KwResult{Answer $ans$.}
        \textbf{Initialize:}
        $\rm{KAMAC} \leftarrow []$.\\
        \Comment{Define prompts. More details in all prompts refer to Appendix \ref{sec: prompt}.}
        $r \leftarrow 1$, $Consensus \leftarrow$False, $KG \leftarrow$False.    \\
        $P_1$: Expert Recruitment Prompt; $P_2$: Initial Assessment Prompt. \\
        $P_3$: Agent Interaction Prompt; 
        $P_4$: KG Prompt for Recruited Experts. \\ $P_5$: KG Prompt for Expert Recruitment;
        $P_6$: Agent Update Prompt. \\
        $P_7$: Final Decision Prompt. \\
        \BlankLine
        \vspace{-3pt}
        \Comment{\textbf{Initial consultation.}}
        \Comment{Recruit $N$ clinician agents.}
        ($Agent_1$, $Agent_2$, $\cdots$, $Agent_N$) $\leftarrow$ Recruit($\mathcal{Q}$, $\rm{KAMAC}$, $P_1$) \\
        \Comment{Clinician agents consist of a multi-disciplinary team.}
        $\rm{KAMAC}$ $\leftarrow$ ($Agent_1$, $Agent_2$, $\cdots$, $Agent_N$) \\
        \Comment{Initial assessment.}
        ($Option_1$, $Option_2$, $\cdots$, $Option_N$)
        $\leftarrow$ Chat($\mathcal{Q}$, $\rm{KAMAC}$, $P_2$) \\
        \Comment{Concat all options as feedback.}
        $Feedback$ $\leftarrow$ Concat($Option_1$, $Option_2$, $\cdots$, $Option_N$) \\
        \Comment{\textbf{Knowledge-driven collaborative discussion.}}
        \While{$r \leq R$, \rm{and} $not$ $Consensus$, \rm{and} $not$ $KG$}{
            \Comment{Exchange agent's comments and determine the consensus.}
            $Consensus$, $Feedback$ $\leftarrow$ Chat($\mathcal{Q}$, $\rm{KAMAC}$, $Feedback$, $P_3$) \\
            \Comment{Assess whether any additional specialist is needed to fill a knowledge or diagnostic gap.}
            $KG$ $\leftarrow$ Chat($\mathcal{Q}$, $\rm{KAMAC}$, $Feedback$, $P_4$) \\
            \If {$KG$}
                {
                    \Comment{Expert recruitment for recruiting additional experts during the discussion.}
                    ($Agent_{N+1}$, $Agent_{N+2}$, $\cdots$, $Agent_{M}$) $\leftarrow$ Recruit($\mathcal{Q}$, $\rm{KAMAC}$, $P_5$) \\
                    \Comment{Review all options and provide comments as feedback.}
                    ($Option_{N+1}$, $Option_{N+2}$, $\cdots$, $Option_{M}$) $\leftarrow$ Chat(($Agent_{N+1}$, $Agent_{N+2}$, $\cdots$, $Agent_{M}$), $Feedback$, $P_2$) \\
                    $Feedback$ $\leftarrow$ Concat($Feedback$, $Option_{N+1}$, $Option_{N+2}$, $\cdots$, $Option_M$) \\
                    \Comment{Exchange agent's comments and determine the consensus.}
                    $Consensus$, $Feedback$ $\leftarrow$ Chat($\mathcal{Q}$, ($Agent_{N+1}$, $Agent_{N+2}$, $\cdots$, $Agent_{M}$), \\$Feedback$, $P_3$) \\
                    \Comment{Update KAMAC.}
                    $\rm{KAMAC}$ $\leftarrow$ ($\rm{KAMAC}$, $\cdots$, $Agent_{N+1}$, $\cdots$, $Agent_M$) \\
                    $KG \leftarrow$False 
                }
            \Comment{Update agent's comments.}
            $Feedback$
            $\leftarrow$ Chat($\mathcal{Q}$, $\rm{KAMAC}$, $Feedback$, $P_6$) \\
            $r \leftarrow r +1$ \\
            }
            \Comment{\textbf{Make the final decision by LLMs.}}
            $ans \leftarrow$ Moderator($\mathcal{Q}$, $Feedback$, $P_7$) 

            \Return $ans$
\end{algorithm*}

\section{Method}
\label{sec:method}

\subsection{Overview}
Figure~\ref{fig:main_fig} presents the KAMAC framework, which comprises three main stages: (1) Initial Consultation:  KAMAC begins with a single/multiple expert agents, which evaluate the case and provide initial feedback for ongoing discussion; (2) Knowledge-driven Collaborative Discussion: Agents engage in a structured, knowledge-guided dialogue to determine whether further expertise is required and then adaptively expands the team and promotes structured discussions among agents, guided by domain knowledge and the evolving diagnostic context, and (3) Decision Making: A designated moderator coordinates the final decision process by initiating a voting mechanism among agents.
The pseudocode of KAMAC is shown in Algorithm~\ref{alg:kac}. More details in all prompts refer to Appendix \ref{sec: prompt}.

\subsection{Initial Consultation}
Given a clinical problem $\mathcal{Q}$, KAMAC first performs an initial consultation by recruiting one or more expert agents from a predefined expert pool. These agents represent diverse clinical roles (\eg, radiologist, cardiologist) and are selected based on their relevance to the query using an expert recruitment prompt $P_1$. Each recruited agent independently analyzes the problem using an initial assessment prompt $P_2$, producing diagnostic opinions or treatment suggestions.
The individual responses are aggregated into a consolidated feedback signal, which serves as the basis for initiating collaborative discussion in the next stage. This step simulates a typical initial clinical encounter, where specialists offer their perspectives before deliberation begins.

\subsection{Knowledge-driven Collaborative Discussion}
In this stage, KAMAC facilitates multi-round, knowledge-driven discussions among the recruited expert agents. Each round begins with agents exchanging their views based on the evolving shared context. Using the agent interaction prompt $P_3$, they critique each other’s responses, resolve inconsistencies, and collaboratively refine their reasoning and comments.

At the end of each round, the currently assigned experts are prompted to assess whether a knowledge gap (KG) remains—that is, whether their collective expertise is sufficient to fully address the problem. This self-assessment is facilitated by the KG detection prompt $P_4$, which takes as input the current discussion and feedback. If a gap is detected, KAMAC triggers expert recruitment by issuing a targeted recruitment prompt $P_5$, allowing the system to enlist additional domain-specific agents to address the identified deficiency.

The newly recruited agents receive contextual examples (\ie, the current discussion history) as few-shot input and respond to the original question using the assessment prompt $P_2$, conditioned on the ongoing feedback. Their outputs are appended to the current feedback buffer and integrated into the group discussion in the subsequent round. This recursive process allows progressive team expansion, enabling KAMAC to dynamically adapt to the evolving complexity of the diagnostic scenario.

Throughout the discussion, all agents, including the initial and newly recruited ones, update their reasoning using the agent update prompt $P_6$, which ensures alignment with the current collective context. This process continues until either (1) a consensus is reached via iterative agreement checks using $P_3$, or (2) a maximum number of discussion rounds $R$ is reached.

\subsection{Decision Making}
In the final stage, KAMAC invokes a moderator agent, typically a general-purpose LLM, to generate the final decision. The moderator receives the latest set of agent comments and the full discussion history and synthesizes a response via a decision prompt ($P_7$).

\section{Experiments}
\label{sec: exp}
\subsection{Datasets}
%In this work, we perform experiments on two medical datasets, \ie, MedQA and Progn-VQA. 
%We utilize the testing set from MedQA~\cite{medqa} and Progn-VQA~\cite{radcure} datasets to validate the proposed KAMAC framework. All datasets are public and can be downloaded.
To evaluate the proposed KAMAC framework, we conduct experiments on the testing sets of two publicly available medical question answering (QA) datasets: MedQA~\cite{medqa} and Progn-VQA~\cite{radcure}.
\paragraph{MedQA} We use all 1273 samples in the testing set. This dataset describes the United States Medical Licensing Examination and includes questions, multiple-choice questions, and answers.
% \paragraph{Path-VQA} The complete Path-VQA has 3391 samples, including open-ended questions, images, and answers. We refer to the method~\cite{mdagents} experiment settings and select more samples with 600 samples to make the samples more diverse for effectiveness evaluation.
\paragraph{Progn-VQA} We use all 750 Visual Question Answering (VQA) pairs in the testing set. The dataset includes head and neck cancer Computed Tomography (CT) image volumes collected from 2005-2017 treated with definitive radiotherapy at the University Health Network in Toronto, Canada. It also contains the corresponding regions of interest (ROIs) and structured patient information in RTSTRUCT format with standardized descriptions, including demographic, clinical, and treatment information based on the 7$th$ edition TNM staging system and AJCC (American Joint Committee on Cancer). The dataset contains patient information, CT image volumes and ROIs, and the patient's survival status at the last follow-up. Please see Table~\ref{tab:variables} for understanding the clinical and imaging information used in the dataset. 
For CT input, we selected the axial slice with the largest cross-sectional area of the ROI. More details on the input clinical and imaging variables are provided in Appendix~\ref{sec: Progn}.

% 对于CT的输入，用最大横截面，more details on input clinical and imaging variables refer to Appendix \ref{sec: prompt}.}

\begin{table*}[ht!]
\footnotesize
\centering
\caption{Main results on four common metrics across MedQA and Progn-VQA datasets, evaluated using GPT-4.1-mini. 
\textbf{Bold} values indicate the best performance.
% , and \underline{underlined} values indicate the second-best.
%Bold values indicate the best performance, and underlined values indicate the second-best. 
Here, `SA' means the single-agent methods, and `MA' means the multi-agent methods. Gray-highlighted cells indicate the average score.}
\begin{adjustbox}{width=\textwidth}
\begin{tabular}{llcccclccccl}
\toprule
\multirow{2}{*}{\textbf{Methods}} & \multirow{2}{*}{\textbf{Types}} & \multicolumn{4}{c}{\textbf{MedQA}} & \multirow{2}{*}{\textbf{Avg}} &  \multicolumn{4}{c}{\textbf{Progn-VQA}} & \multirow{2}{*}{\textbf{Avg}} \\
& & Acc & Prec & Spec & Recall & &  Acc  & Prec  & Spec  & Recall &  \\
\midrule
Single-agent & SA & 79.50 & 79.65 & 94.86 &  79.36 & \cellcolor{gray!20}83.34 &  86.00 & 86.28 & 14.79 & 97.21 & \cellcolor{gray!20}71.07 \\
+CoT         & SA & 84.21  & 84.82  & 96.03  &  84.02 & \cellcolor{gray!20}87.27 & 84.67 & 86.29 & 15.52 & 97.32 & \cellcolor{gray!20}70.95 \\
\midrule
Majority Voting  & MA  &  86.49 & 86.93  & 96.60  &   86.38 & \cellcolor{gray!20}89.10  &   86.27   &   86.12   &  12.17 & \textbf{99.84}  &  \cellcolor{gray!20}71.10 \\
Consensus & MA & 80.68  & 80.70  & 95.15 &  80.59 & \cellcolor{gray!20}84.28  &   86.86   &  86.81    &   31.85   &  \uline{98.86}  & \cellcolor{gray!20}76.09 \\
MDAgents     & MA & \uline{87.74} & \uline{87.92} & \uline{96.92} & \uline{87.55} & \cellcolor{gray!20}90.03 & \uline{87.01} & \uline{88.83} & \uline{33.70} & 96.21 & \cellcolor{gray!20}76.44 \\
KAMAC    & MA &  \textbf{88.14} & \textbf{88.30}  &  \textbf{97.02} &  \textbf{88.11} & \cellcolor{gray!20}\textbf{90.39} & \textbf{87.20} & \textbf{89.79} & \textbf{40.52} & 95.74 &  \cellcolor{gray!20}\textbf{78.31} \\
\bottomrule
\end{tabular}
\end{adjustbox}
\label{tab:main-results}
\end{table*}

\begin{table*}[t!]
\centering
\caption{Performance comparison of Baseline and KAMAC on MedQA and Progn-VQA using DeepSeek-R1 and GPT-4.1-mini across four metrics and their average. Gray-highlighted cells indicate the average score, with relative improvements shown in small colored text. Where `Baseline' means single-agent+CoT.}
\begin{adjustbox}{width=\textwidth}
\begin{tabular}{l ccccc ccccc}
\toprule
\multirow{2}{*}{\textbf{Method}} & \multicolumn{4}{c}{\textbf{MedQA}} & \multirow{2}{*}{\textbf{Avg}} & \multicolumn{4}{c}{\textbf{Progn-VQA}} & \multirow{2}{*}{\textbf{Avg}} \\
& Acc & Prec & Spec & Recall & & Acc & Prec & Spec & Recall & \\
\midrule
Baseline (DeepSeek-R1) & 88.14  & 88.12  & \uline{97.03}  &  88.00 & \cellcolor{gray!20}90.32 & 77.87 & 88.11 & 37.07 & 85.33    & \cellcolor{gray!20}72.10  \\
KAMAC (DeepSeek-R1) & \textbf{89.63} & \textbf{89.53} & \textbf{97.41} & \textbf{89.50} & \cellcolor{gray!20}\textbf{91.52}{\scriptsize\textcolor{green!50!black}{(+1.20)}} & \uline{86.13} & \uline{88.41} & \uline{31.03} & \uline{96.21} & \cellcolor{gray!20}\uline{75.45} {\scriptsize\textcolor{green!50!black}{(+3.35)}} \\
\midrule
Baseline (GPT-4.1-mini) & 84.21  & 84.82  & 96.03  &  84.02 & \cellcolor{gray!20}87.27 & 84.67 & 86.29 & 15.52 & \textbf{97.32} & \cellcolor{gray!20}70.95 \\
KAMAC (GPT-4.1-mini) & \uline{88.14} & \uline{88.30} & 97.02 & \uline{88.11} & \cellcolor{gray!20}\uline{90.39} {\scriptsize\textcolor{green!50!black}{(+3.12)}} & \textbf{87.20} & \textbf{89.79} & \textbf{40.52} & 95.74 & \cellcolor{gray!20}\textbf{78.31} {\scriptsize\textcolor{green!50!black}{(+7.36)}} \\
\bottomrule
\end{tabular}
\end{adjustbox}
\label{tab:comp-ds}
\end{table*}

\subsection{Implementation Details}
%We use GPT-4.1-mini for main experiments and set the temperature to 0. We also compare the proposed methods with the DeepSeek-R1~\cite{deepseek-r1} model in Table~\ref{tab:comp-ds}. Then, we store each chat record into a local file. For the same question, we read the local file, ensuring the same initial medical comments are generated by the same role before the discussion starts, and then reopen the discussion. In the end, the final decision result will only depend on the proposed collaboration method. The maximum number of discussions $R$ is set to 3. The reason why we chose GPT-4.1-mini is that it delivers near state-of-the-art medical reasoning with minimal latency, predictable computational cost, and fully deterministic outputs—benefits that larger models like GPT-4 or retrieval-optimized variants such as DeepSeek-R1 cannot offer simultaneously without incurring higher overhead or output variability.
%
We use GPT-4.1-mini~\footnote{https://openai.com/index/gpt-4-1/} as the primary model for all experiments, with the temperature set to 0 to ensure deterministic outputs. In addition, we compare our proposed method with DeepSeek-R1~\cite{deepseek-r1}, as shown in Table~\ref{tab:comp-ds}. For each medical question, we store the corresponding chat history in a local file. When revisiting the same question, the system loads the saved file to regenerate consistent initial medical comments from each role before resuming the collaborative discussion. The final decision is made solely based on the proposed collaboration method. The maximum number of discussion rounds $R$ is set to 3. The initial number of experts is set to 1.
We select GPT-4.1-mini due to its strong medical reasoning capabilities, low latency, predictable computational cost, and fully deterministic behavior. These advantages make it preferable for our controlled evaluation setting, in contrast to larger models such as GPT-4 or retrieval-enhanced models like DeepSeek-R1, which often entail higher overhead and less consistent outputs.

\subsubsection{Comparison Methods}

The compared methods include: (1) \textbf{Single-agent}, which uses an LLM for decision-making, where the question and the answer template are input to output an answer, (2) \textbf{Chain of Thought} (CoT)~\cite{cot}, which combines the single-agent backbone with a step-by-step prompt to conduct analysis and decision-making, 
% (3) \textbf{Few-shot}~\cite{fewshot}, which randomly samples different questions with the correct answers as exemplars fed into the single-agent backbone to enhance LLM reasoning, 
(3) \textbf{Majority Voting}, which is used in multi-agent decision-making methods~\cite{reconcile, yang2024llm, voting} for making final decision with more that 50\% votes. (4) \textbf{Consensus}, which is also adopted in~\cite{voting, mdagents}. 
% (5) \xie{Observation-driven MAC-}\textbf{AutoAgents}~\cite{autoagents} for making final decision by prompting the agents to review all answers. 
(5) \textbf{MDAgents}~\cite{mdagents} is an advanced multi-agent framework that performs problem-driven expert recruitment, MAC, and consensus decision to output the final results.

\begin{figure}[ht!]
    \centering
    \begin{tabular}{c}
         \includegraphics[width=0.96\linewidth]{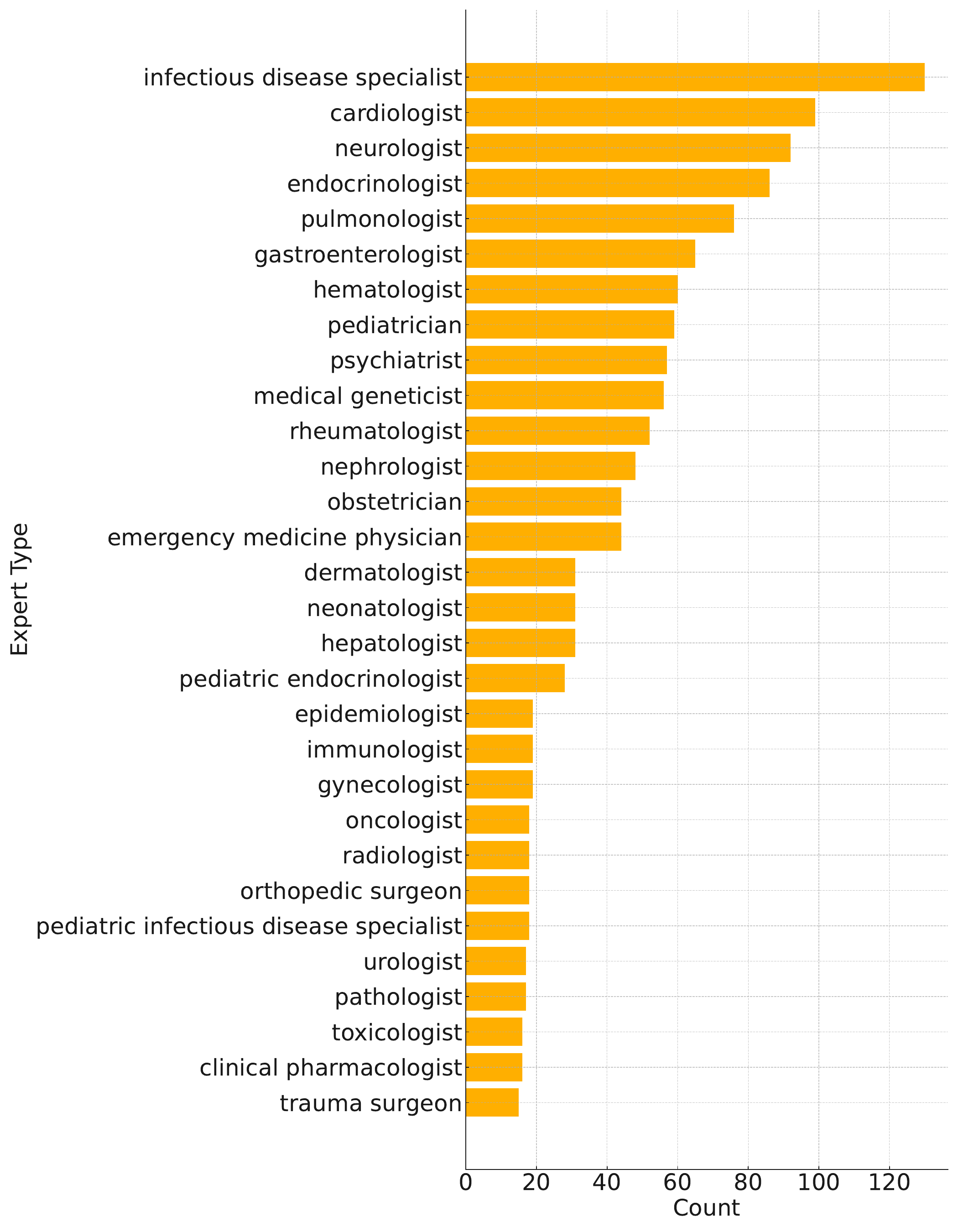} \\
         (a) Number of Initial Experts: 1 \\
         \includegraphics[width=0.96\linewidth]{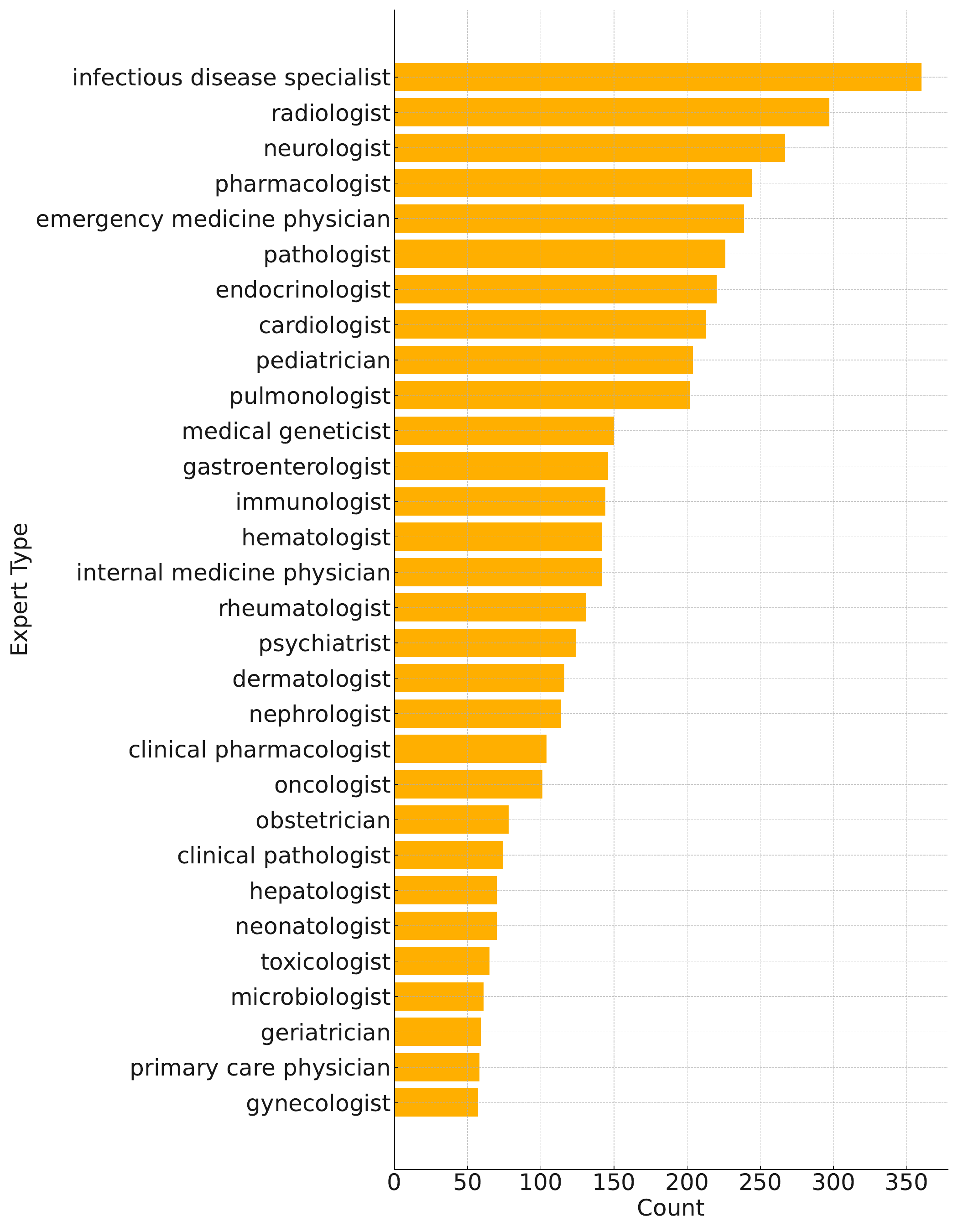} \\
         (b) Number of Initial Experts: 5
    \end{tabular}
    
    \caption{Histogram illustrating the impact of initial expert settings on the final top-30 expert distribution in our method on the MedQA dataset. "Count" denotes the total frequency of each expert type. An 80\% overlap in expert types is observed between the 1- and 5-expert settings.}
    \label{fig:hist}
\end{figure}

\begin{table*}[ht!]
\centering
\caption{Discussion on the number of initial agents on the MedQA and Progn-VQA datasets. Gray-highlighted cells indicate the average score.}
\begin{adjustbox}{width=0.85\textwidth}
\begin{tabular}{cccccc ccccc}
\toprule
\multirow{1}{*}{\textbf{Initial Agents}} & \multicolumn{4}{c}{\textbf{MedQA}} & \multirow{2}{*}{\textbf{Avg}} & \multicolumn{4}{c}{\textbf{Progn-VQA}} & \multirow{2}{*}{\textbf{Avg}} \\
\textbf{Number}& Acc & Prec & Spec & Recall & & Acc & Prec & Spec & Recall & \\
\midrule
1              &  88.14 & 88.30  &  97.02 &  88.11 & \cellcolor{gray!20}90.39  & 87.20 & 89.79 & 40.52 & 95.74 & \cellcolor{gray!20}78.31 \\
3              &  87.98    &  88.16   &   96.98   &   87.82  &  \cellcolor{gray!20}90.24 &   86.40  &  90.43   &   45.69  &  93.85  &  \cellcolor{gray!20}79.09 \\
5              &  80.28   &   80.31   &   95.06   &   80.13  & \cellcolor{gray!20}83.95 &  89.10 & 89.54 & 35.43 & 96.69 & \cellcolor{gray!20}77.69 \\
\bottomrule
\end{tabular}
\label{tab:num-agents}
\end{adjustbox}
\end{table*}

\subsection{Evaluation Metrics}
We evaluate the proposed method using four standard metrics: accuracy (Acc), precision (Prec), specificity (Spec), and recall score (Recall).

\section{Results and Analysis}
\subsection{Comparisons with State-of-the-Arts}
In Table~\ref{tab:main-results}, the proposed method achieves improved results on four metrics compared to multiple methods on the MedQA dataset. For the Progn-VQA dataset, the proposed method achieves better results on the Acc, Prec, and Spec metrics. In addition,  KAMAC leverages knowledge-driven prompts to detect KG and expand experts to form multi-agent collaborative discussions. Focusing on multi-agent-based methods, both the majority voting and consensus are set to five experts, while MDAgents adopts a single agent, a multi-disciplinary team with five experts, and an integrated care team with nine experts. Compared with them, the proposed method can achieve better results, which demonstrates that our method overcomes the limitation of knowledge in the single-agent model and has a more suitable multi-disciplinary team to enhance multi-agent reasoning and collaboration.

In Table~\ref{tab:comp-ds}, we further evaluate our method on another model, DeepSeek-R1. In our method, the initial number of experts is set to 1, which is consistent with the baseline method (single-agent + CoT), but the experimental results are better than the baseline method. This improvement shows that our method can be generalized to more LLM models. In addition, 
this improvement aligns with the actual clinical treatment scenario, where clinical treatment allows the dynamic addition of experts according to the patient's clinical treatment situation, thereby carrying out more effective treatment.
It contributes to optimizing the best treatment options and limited clinical resources in medical applications.

\subsection{Discussion of KAMAC}

\paragraph{Results of Different Multi-agent Methods}
\xiao{In Table~\ref{tab:main-results}, we present different multi-agent methods. Our approach outperforms the static multi-agent methods (majority voting and Consensus) and the MDAgent method on both datasets. In the static multi-agent methods, five experts were set to conduct a multi-agent collaboration. In MDAgents, there were 2.41 and 4.34 experts per case. Compared to the above methods, KAMAC recruited 1.28 and 2.41 experts per case, respectively, which is still significantly more efficient than the full multi-agent approach and is 53\% and 56\% lower than MDAgents, respectively. In addition, we also perform statistical analysis and cost-efficiency in Appendix~\ref{sec: more_discussion}.}

\paragraph{Comparison of Initial Number of Experts}
\xiao{As shown in Table~\ref{tab:num-agents}, using one initial expert outperforms other configurations. This indicates that starting with a single agent promotes more targeted recruitment. With only one initial agent, KAC-MAF more accurately identifies knowledge gaps and recruits the most relevant experts. In contrast, beginning with five agents may introduce overlapping or irrelevant perspectives early, increasing redundancy and noise in subsequent decisions. Thus, fewer initial experts allow for more precise adaptation and reduce early overfitting.} In Figure~\ref{fig:hist}, we show the expert distribution for KAMAC with 1 versus 5 initial experts. Among the top 30 experts, 24 are common to both settings (80\% overlap). However, the total number of experts remains lower when starting with one expert compared to five.

\section{Conclusion}
This work presents KAMAC, a knowledge-driven adaptive multi-agent collaboration framework that brings structured, dynamic reasoning into medical decision-making with LLMs. By allowing agents to actively assess their own limitations and request additional expertise when needed, KAMAC overcomes the rigidity of traditional multi-agent setups and more faithfully mirrors real-world clinical workflows. Our experiments on two real-world medical QA benchmarks demonstrate that KAMAC consistently outperforms both single-agent and existing multi-agent baselines.

Beyond accuracy improvements, KAMAC offers deeper insights into AI collaboration: decision quality improves not merely through more parameters or agents, but through adaptive, feedback-driven interaction grounded in knowledge awareness. This framework brings multi-agent LLM systems closer to real-world clinical workflows, where expert composition evolves with case complexity.
Future directions include modeling agent uncertainty and integrating clinician-in-the-loop feedback to further support real-time deployment in medical environments.

\section*{Limitations}
\label{sec: limit}

While KAMAC demonstrates promising results, it has several limitations. The current framework focuses on textual and imaging inputs; future work could incorporate additional modalities such as genomic or longitudinal clinical data to support a wider range of medical tasks. Although KAMAC achieves strong performance without fine-tuning the underlying LLMs, domain-specific fine-tuning may further improve accuracy and agent-role fidelity. However, this would introduce significant computational overhead and is challenged by the scarcity of high-quality, labeled medical data. Balancing accuracy gains with efficiency and data availability remains an important direction for future fine-tuning efforts. 

\section*{Acknowledgments}
This work was supported in part by Natural Science Foundation of Sichuan Province (2024NSFSC0038) and the NSFC (12171072).

% This document has been adapted
% by Steven Bethard, Ryan Cotterell and Rui Yan
% from the instructions for earlier ACL and NAACL proceedings, including those for
% ACL 2019 by Douwe Kiela and Ivan Vuli\'{c},
% NAACL 2019 by Stephanie Lukin and Alla Roskovskaya,
% ACL 2018 by Shay Cohen, Kevin Gimpel, and Wei Lu,
% NAACL 2018 by Margaret Mitchell and Stephanie Lukin,
% Bib\TeX{} suggestions for (NA)ACL 2017/2018 from Jason Eisner,
% ACL 2017 by Dan Gildea and Min-Yen Kan,
% NAACL 2017 by Margaret Mitchell,
% ACL 2012 by Maggie Li and Michael White,
% ACL 2010 by Jing-Shin Chang and Philipp Koehn,
% ACL 2008 by Johanna D. Moore, Simone Teufel, James Allan, and Sadaoki Furui,
% ACL 2005 by Hwee Tou Ng and Kemal Oflazer,
% ACL 2002 by Eugene Charniak and Dekang Lin,
% and earlier ACL and EACL formats written by several people, including
% John Chen, Henry S. Thompson and Donald Walker.
% Additional elements were taken from the formatting instructions of the \emph{International Joint Conference on Artificial Intelligence} and the \emph{Conference on Computer Vision and Pattern Recognition}.

% Bibliography entries for the entire Anthology, followed by custom entries
%\bibliography{anthology,custom}
% Custom bibliography entries only
\bibliography{acl_latex}

\begin{thebibliography}{33}
\providecommand{\natexlab}[1]{#1}

\bibitem[{Abdelnabi et~al.(2024)Abdelnabi, Gomaa, Sivaprasad, Sch{\"o}nherr, and Fritz}]{cooperation}
Sahar Abdelnabi, Amr Gomaa, Sarath Sivaprasad, Lea Sch{\"o}nherr, and Mario Fritz. 2024.
\newblock Cooperation, competition, and maliciousness: Llm-stakeholders interactive negotiation.
\newblock \emph{Advances in Neural Information Processing Systems}, 37:83548--83599.

\bibitem[{Achiam et~al.(2023)Achiam, Adler, Agarwal, Ahmad, Akkaya, Aleman, Almeida, Altenschmidt, Altman, Anadkat et~al.}]{gpt}
Josh Achiam, Steven Adler, Sandhini Agarwal, Lama Ahmad, Ilge Akkaya, Florencia~Leoni Aleman, Diogo Almeida, Janko Altenschmidt, Sam Altman, Shyamal Anadkat, and 1 others. 2023.
\newblock Gpt-4 technical report.
\newblock \emph{arXiv preprint arXiv:2303.08774}.

\bibitem[{Chan et~al.(2023)Chan, Chen, Su, Yu, Xue, Zhang, Fu, and Liu}]{chateval}
Chi-Min Chan, Weize Chen, Yusheng Su, Jianxuan Yu, Wei Xue, Shanghang Zhang, Jie Fu, and Zhiyuan Liu. 2023.
\newblock Chateval: Towards better llm-based evaluators through multi-agent debate.
\newblock \emph{arXiv preprint arXiv:2308.07201}.

\bibitem[{Chen et~al.(2023{\natexlab{a}})Chen, Dong, Shu, Zhang, Sesay, Karlsson, Fu, and Shi}]{autoagents}
Guangyao Chen, Siwei Dong, Yu~Shu, Ge~Zhang, Jaward Sesay, B{\"o}rje~F Karlsson, Jie Fu, and Yemin Shi. 2023{\natexlab{a}}.
\newblock Autoagents: A framework for automatic agent generation.
\newblock \emph{arXiv preprint arXiv:2309.17288}.

\bibitem[{Chen et~al.(2023{\natexlab{b}})Chen, Saha, and Bansal}]{reconcile}
Justin Chih-Yao Chen, Swarnadeep Saha, and Mohit Bansal. 2023{\natexlab{b}}.
\newblock Reconcile: Round-table conference improves reasoning via consensus among diverse llms.
\newblock \emph{arXiv preprint arXiv:2309.13007}.

\bibitem[{Chen et~al.(2025)Chen, Li, Yang, Wang, Dong, and Gao}]{mdteamgpt}
Kai Chen, Xinfeng Li, Tianpei Yang, Hewei Wang, Wei Dong, and Yang Gao. 2025.
\newblock Mdteamgpt: A self-evolving llm-based multi-agent framework for multi-disciplinary team medical consultation.
\newblock \emph{arXiv preprint arXiv:2503.13856}.

\bibitem[{Chen et~al.(2023{\natexlab{c}})Chen, Su, Zuo, Yang, Yuan, Qian, Chan, Qin, Lu, Xie et~al.}]{agentverse}
Weize Chen, Yusheng Su, Jingwei Zuo, Cheng Yang, Chenfei Yuan, Chen Qian, Chi-Min Chan, Yujia Qin, Yaxi Lu, Ruobing Xie, and 1 others. 2023{\natexlab{c}}.
\newblock Agentverse: Facilitating multi-agent collaboration and exploring emergent behaviors in agents.
\newblock \emph{arXiv preprint arXiv:2308.10848}, 2(4):6.

\bibitem[{Guo et~al.(2025)Guo, Yang, Zhang, Song, Zhang, Xu, Zhu, Ma, Wang, Bi et~al.}]{deepseek-r1}
Daya Guo, Dejian Yang, Haowei Zhang, Junxiao Song, Ruoyu Zhang, Runxin Xu, Qihao Zhu, Shirong Ma, Peiyi Wang, Xiao Bi, and 1 others. 2025.
\newblock Deepseek-r1: Incentivizing reasoning capability in llms via reinforcement learning.
\newblock \emph{arXiv preprint arXiv:2501.12948}.

\bibitem[{Hyland et~al.(2023)Hyland, Bannur, Bouzid, Castro, Ranjit, Schwaighofer, P{\'e}rez-Garc{\'\i}a, Salvatelli, Srivastav, Thieme et~al.}]{maira}
Stephanie~L Hyland, Shruthi Bannur, Kenza Bouzid, Daniel~C Castro, Mercy Ranjit, Anton Schwaighofer, Fernando P{\'e}rez-Garc{\'\i}a, Valentina Salvatelli, Shaury Srivastav, Anja Thieme, and 1 others. 2023.
\newblock Maira-1: A specialised large multimodal model for radiology report generation.
\newblock \emph{arXiv preprint arXiv:2311.13668}.

\bibitem[{Jin et~al.(2021)Jin, Pan, Oufattole, Weng, Fang, and Szolovits}]{medqa}
Di~Jin, Eileen Pan, Nassim Oufattole, Wei-Hung Weng, Hanyi Fang, and Peter Szolovits. 2021.
\newblock What disease does this patient have? a large-scale open domain question answering dataset from medical exams.
\newblock \emph{Applied Sciences}, 11(14):6421.

\bibitem[{Kaesberg et~al.(2025)Kaesberg, Becker, Wahle, Ruas, and Gipp}]{voting}
Lars~Benedikt Kaesberg, Jonas Becker, Jan~Philip Wahle, Terry Ruas, and Bela Gipp. 2025.
\newblock Voting or consensus? decision-making in multi-agent debate.
\newblock \emph{arXiv preprint arXiv:2502.19130}.

\bibitem[{Kim et~al.(2024)Kim, Park, Jeong, Chan, Xu, McDuff, Lee, Ghassemi, Breazeal, and Park}]{mdagents}
Yubin Kim, Chanwoo Park, Hyewon Jeong, Yik~S Chan, Xuhai Xu, Daniel McDuff, Hyeonhoon Lee, Marzyeh Ghassemi, Cynthia Breazeal, and Hae~W Park. 2024.
\newblock Mdagents: An adaptive collaboration of llms for medical decision-making.
\newblock \emph{Advances in Neural Information Processing Systems}, 37:79410--79452.

\bibitem[{Kodner and Spreeuwenberg(2002)}]{kodner2002integrated}
Dennis~L Kodner and Cor Spreeuwenberg. 2002.
\newblock Integrated care: meaning, logic, applications, and implications--a discussion paper.
\newblock \emph{International journal of integrated care}, 2:e12.

\bibitem[{Li et~al.(2024{\natexlab{a}})Li, Zhang, Yu, Fu, and Ye}]{AgentForest}
Junyou Li, Qin Zhang, Yangbin Yu, Qiang Fu, and Deheng Ye. 2024{\natexlab{a}}.
\newblock More agents is all you need.
\newblock \emph{Transactions on Machine Learning Research}.

\bibitem[{Li et~al.(2024{\natexlab{b}})Li, Balachandran, Feng, Ilgen, Pierson, Koh, and Tsvetkov}]{mediq}
Stella Li, Vidhisha Balachandran, Shangbin Feng, Jonathan Ilgen, Emma Pierson, Pang Wei~W Koh, and Yulia Tsvetkov. 2024{\natexlab{b}}.
\newblock Mediq: Question-asking llms and a benchmark for reliable interactive clinical reasoning.
\newblock \emph{Advances in Neural Information Processing Systems}, 37:28858--28888.

\bibitem[{Liang et~al.(2023)Liang, He, Jiao, Wang, Wang, Wang, Yang, Shi, and Tu}]{encouraging}
Tian Liang, Zhiwei He, Wenxiang Jiao, Xing Wang, Yan Wang, Rui Wang, Yujiu Yang, Shuming Shi, and Zhaopeng Tu. 2023.
\newblock Encouraging divergent thinking in large language models through multi-agent debate.
\newblock \emph{arXiv preprint arXiv:2305.19118}.

\bibitem[{Liu et~al.(2024)Liu, Feng, Xue, Wang, Wu, Lu, Zhao, Deng, Zhang, Ruan et~al.}]{deepseek-v3}
Aixin Liu, Bei Feng, Bing Xue, Bingxuan Wang, Bochao Wu, Chengda Lu, Chenggang Zhao, Chengqi Deng, Chenyu Zhang, Chong Ruan, and 1 others. 2024.
\newblock Deepseek-v3 technical report.
\newblock \emph{arXiv preprint arXiv:2412.19437}.

\bibitem[{Nori et~al.(2023)Nori, King, McKinney, Carignan, and Horvitz}]{gpt-medical}
Harsha Nori, Nicholas King, Scott~Mayer McKinney, Dean Carignan, and Eric Horvitz. 2023.
\newblock Capabilities of gpt-4 on medical challenge problems.
\newblock \emph{arXiv preprint arXiv:2303.13375}.

\bibitem[{Singhal et~al.(2022)Singhal, Azizi, Tu, Mahdavi, Wei, Chung, Scales, Tanwani, Cole-Lewis, Pfohl et~al.}]{ER}
Karan Singhal, Shekoofeh Azizi, Tao Tu, S~Sara Mahdavi, Jason Wei, Hyung~Won Chung, Nathan Scales, Ajay Tanwani, Heather Cole-Lewis, Stephen Pfohl, and 1 others. 2022.
\newblock Large language models encode clinical knowledge.
\newblock \emph{arXiv preprint arXiv:2212.13138}.

\bibitem[{Sutton et~al.(2020)Sutton, Pincock, Baumgart, Sadowski, Fedorak, and Kroeker}]{sutton2020overview}
Reed~T Sutton, David Pincock, Daniel~C Baumgart, Daniel~C Sadowski, Richard~N Fedorak, and Karen~I Kroeker. 2020.
\newblock An overview of clinical decision support systems: benefits, risks, and strategies for success.
\newblock \emph{NPJ digital medicine}, 3(1):17.

\bibitem[{Tang et~al.(2023)Tang, Zou, Zhang, Li, Zhao, Zhang, Cohan, and Gerstein}]{medagents}
Xiangru Tang, Anni Zou, Zhuosheng Zhang, Ziming Li, Yilun Zhao, Xingyao Zhang, Arman Cohan, and Mark Gerstein. 2023.
\newblock Medagents: Large language models as collaborators for zero-shot medical reasoning.
\newblock \emph{arXiv preprint arXiv:2311.10537}.

\bibitem[{Team et~al.(2024)Team, Georgiev, Lei, Burnell, Bai, Gulati, Tanzer, Vincent, Pan, Wang et~al.}]{gemini}
Gemini Team, Petko Georgiev, Ving~Ian Lei, Ryan Burnell, Libin Bai, Anmol Gulati, Garrett Tanzer, Damien Vincent, Zhufeng Pan, Shibo Wang, and 1 others. 2024.
\newblock Gemini 1.5: Unlocking multimodal understanding across millions of tokens of context.
\newblock \emph{arXiv preprint arXiv:2403.05530}.

\bibitem[{Thawakar et~al.(2024)Thawakar, Shaker, Mullappilly, Cholakkal, Anwer, Khan, Laaksonen, and Khan}]{xraygpt}
Omkar~Chakradhar Thawakar, Abdelrahman~M Shaker, Sahal~Shaji Mullappilly, Hisham Cholakkal, Rao~Muhammad Anwer, Salman Khan, Jorma Laaksonen, and Fahad Khan. 2024.
\newblock Xraygpt: Chest radiographs summarization using large medical vision-language models.
\newblock In \emph{Proceedings of the 23rd workshop on biomedical natural language processing}, pages 440--448.

\bibitem[{Tran et~al.(2025)Tran, Dao, Nguyen, Pham, O'Sullivan, and Nguyen}]{mc_survey}
Khanh-Tung Tran, Dung Dao, Minh-Duong Nguyen, Quoc-Viet Pham, Barry O'Sullivan, and Hoang~D Nguyen. 2025.
\newblock Multi-agent collaboration mechanisms: A survey of llms.
\newblock \emph{arXiv preprint arXiv:2501.06322}.

\bibitem[{Wang et~al.(2024)Wang, Wang, Athiwaratkun, Zhang, and Zou}]{moa}
Junlin Wang, Jue Wang, Ben Athiwaratkun, Ce~Zhang, and James Zou. 2024.
\newblock Mixture-of-agents enhances large language model capabilities.
\newblock \emph{arXiv preprint arXiv:2406.04692}.

\bibitem[{Wei et~al.(2022)Wei, Wang, Schuurmans, Bosma, Xia, Chi, Le, Zhou et~al.}]{cot}
Jason Wei, Xuezhi Wang, Dale Schuurmans, Maarten Bosma, Fei Xia, Ed~Chi, Quoc~V Le, Denny Zhou, and 1 others. 2022.
\newblock Chain-of-thought prompting elicits reasoning in large language models.
\newblock \emph{Advances in neural information processing systems}, 35:24824--24837.

\bibitem[{Welch et~al.(2023)Welch, Kim, Hope, Huang, Lu, Marsilla, Kazmierski, Rey-McIntyre, Patel, O’Sullivan et~al.}]{radcure}
ML~Welch, S~Kim, A~Hope, SH~Huang, Z~Lu, J~Marsilla, M~Kazmierski, K~Rey-McIntyre, T~Patel, B~O’Sullivan, and 1 others. 2023.
\newblock Computed tomography images from large head and neck cohort (radcure).
\newblock \emph{The Cancer Imaging Archive}.

\bibitem[{Xiong et~al.(2024)Xiong, Jin, Lu, and Zhang}]{medrag}
Guangzhi Xiong, Qiao Jin, Zhiyong Lu, and Aidong Zhang. 2024.
\newblock Benchmarking retrieval-augmented generation for medicine.
\newblock In \emph{Findings of the Association for Computational Linguistics ACL 2024}, pages 6233--6251.

\bibitem[{Yang et~al.(2024{\natexlab{a}})Yang, Dalisan, Korecki, Hausladen, and Helbing}]{yang2024llm}
Joshua~C Yang, Damian Dalisan, Marcin Korecki, Carina~I Hausladen, and Dirk Helbing. 2024{\natexlab{a}}.
\newblock Llm voting: Human choices and ai collective decision-making.
\newblock In \emph{Proceedings of the AAAI/ACM Conference on AI, Ethics, and Society}, volume~7, pages 1696--1708.

\bibitem[{Yang et~al.(2024{\natexlab{b}})Yang, Zhang, Wang, Xu, Lin, and Sui}]{critique}
Zhe Yang, Yichang Zhang, Yudong Wang, Ziyao Xu, Junyang Lin, and Zhifang Sui. 2024{\natexlab{b}}.
\newblock Confidence vs critique: A decomposition of self-correction capability for llms.
\newblock \emph{arXiv preprint arXiv:2412.19513}.

\bibitem[{Yuan et~al.(2024)Yuan, Song, Chen, Tan, Li, and Yang}]{yuan2024evoagent}
Siyu Yuan, Kaitao Song, Jiangjie Chen, Xu~Tan, Dongsheng Li, and Deqing Yang. 2024.
\newblock Evoagent: Towards automatic multi-agent generation via evolutionary algorithms.
\newblock In \emph{NeurIPS Workshop on Open-World Agents}.

\bibitem[{Yue et~al.(2024)Yue, Xing, Chen, and Fu}]{clinicalagent}
Ling Yue, Sixue Xing, Jintai Chen, and Tianfan Fu. 2024.
\newblock Clinicalagent: Clinical trial multi-agent system with large language model-based reasoning.
\newblock In \emph{Proceedings of the 15th ACM International Conference on Bioinformatics, Computational Biology and Health Informatics}, pages 1--10.

\bibitem[{Zhang et~al.(2024)Zhang, Zhou, Adhikarla, Yan, Liu, Yu, Liu, Chen, Davison, Ren et~al.}]{generalist}
Kai Zhang, Rong Zhou, Eashan Adhikarla, Zhiling Yan, Yixin Liu, Jun Yu, Zhengliang Liu, Xun Chen, Brian~D Davison, Hui Ren, and 1 others. 2024.
\newblock A generalist vision--language foundation model for diverse biomedical tasks.
\newblock \emph{Nature Medicine}, pages 1--13.

\end{thebibliography}

\newpage
\appendix

% \section{Example Appendix}
\label{sec:appendix}
\section{Clinical Impact and Challenges}
\xiao{In Section~\ref{sec: limit}, we discussed the limitations of KAMAC. Now, we will explore the clinical implications and challenges of multi-agent collaboration.}
\paragraph{Adaptive Multi-agent Collaboration}
\xiao{Clinical decision-making often involves the emergence of new information and continuous refinement of diagnoses. The dynamic multi-agent framework is well-suited to such scenarios, as it enables progressive deepening of the diagnostic process and allows selective inclusion of additional experts when necessary. This flexibility helps ensure that complex or ambiguous cases receive input from the most relevant specialists, thereby improving diagnostic accuracy and comprehensiveness. However, given the inherent uncertainties associated with large language models, the dynamics and stability of the system in real-world clinical applications warrant further investigation in future studies.}

\paragraph{Consensus and Deliberation}
\xiao{In clinical practice, the effective integration of diverse expert opinions is essential. Our framework facilitates progressive discussions by starting with a single agent, allowing the reasoning process to unfold incrementally and making disagreements or uncertainties more explicit. This structure encourages more thorough deliberation. It should be noted that we did not attempt to force consensus among experts; rather, additional experts were introduced selectively to ensure that the collective recommendations are as comprehensive and well-supported as possible.}

\section{More Discussion of KAMAC}
\label{sec: more_discussion}

\begin{table*}[thbp]
\centering
\caption{Discussion on the consensus strategy of the KAMAC method on the MedQA and Progn-VQA datasets. Gray-highlighted cells indicate the average score.}
\begin{adjustbox}{width=1.0\textwidth}
\begin{tabular}{cccccc ccccc}
\toprule
\multirow{2}{*}{\textbf{Consensus Strategy}} & \multicolumn{4}{c}{\textbf{MedQA}} & \multirow{2}{*}{\textbf{Avg}} & \multicolumn{4}{c}{\textbf{Progn-VQA}} & \multirow{2}{*}{\textbf{Avg}} \\
& Acc & Prec & Spec & Recall & & Acc & Prec & Spec & Recall & \\
\midrule
Ensemble Refinement &  87.35 & 87.40  &  96.83 &  87.23 & \cellcolor{gray!20}89.70  & 86.27 & 90.91 & 49.14 & 93.06 & \cellcolor{gray!20}79.85 \\
Majority Voting &  88.14 & 88.30  &  97.02 &  88.11 & \cellcolor{gray!20}90.39  & 87.20 & 89.79 & 40.52 & 95.74 & \cellcolor{gray!20}78.31 \\

\bottomrule
\end{tabular}
\end{adjustbox}
\label{tab: consensus}
\end{table*}

\begin{table*}[thbp]
\centering
\renewcommand{\arraystretch}{1.2}
\caption{Comparison of Baseline, Major Voting, MDAgent, and KAMAC on metrics, expert usage (Number), inference time per case (Time (s)), API calls, and cost on the MedQA dataset.}
\begin{tabular}{lccccccccc}
\hline
\textbf{Method} & \textbf{Acc} & \textbf{Prec}& \textbf{Spec} & \textbf{Recall} & \textbf{Avg} & \textbf{Number}  & \textbf{Time (s)} & \textbf{API Calls} & \textbf{Cost} \\
\hline
Baseline & 84.21  & 84.82  & 96.03  &  84.02 &  87.27 & 1 & 9.56 & 2.20 & 4.01 \\
Majority Voting  &  86.49 & 86.93  & 96.60  &   86.38 & 89.10 & 5 & 39.88 & 12.02 & 8.12  \\
MDAgent  & 87.74 & 87.92 & 96.92 & 87.55 & 90.03 & 2.41 &  15.62  & 3.34 & 6.31 \\
KAMAC  & 88.14 & 88.30  &  97.02 &  88.11 & 90.39 & 1.28 & 10.80 & 2.55 & 5.01 \\
\hline
\end{tabular}
\label{tab: cost}
\end{table*}

\begin{table*}[h]
\centering
\caption{Statistical Significance Comparison of KAMAC and MDAgent vs Baseline: single agent + CoT}
\resizebox{\textwidth}{!}{
\begin{tabular}{llcccccccc}
\toprule
\multicolumn{2}{l}{\textbf{Dataset}} & \multicolumn{4}{c}{KAMAC} & \multicolumn{4}{c}{MDAgent} \\
\cmidrule(lr){3-6}\cmidrule(lr){7-10}
 & Metric & Mean & Std & t-statistic & p-value & Mean & Std & t-statistic & p-value \\
\midrule
\multirow{4}{*}{MedQA}
& Acc 
  & 0.8761   & 0.0016   & 30.07$^{**}$  & 0.0011$^{**}$  
  & 0.8761   & 0.0011   & \textbf{43.33}$^{***}$  & \textbf{0.0005}$^{***}$ \\
& Prec
  & \textbf{0.8765}   & 0.0015   & \textbf{33.05}$^{***}$  & \textbf{0.0009}$^{***}$  
  & 0.8756   & 0.0032   & 15.04$^{**}$  & 0.0044$^{**}$ \\
& Recall       
  & \textbf{0.8748}   & 0.0018   & \textbf{26.15}$^{**}$  & \textbf{0.0015}$^{**}$  
  & 0.8725   & 0.0031   & 13.78$^{**}$  & 0.0052$^{**}$ \\
& Spec
  & 0.9745   & 0.0076   & \textbf{24.74}$^{**}$  & \textbf{0.0016}$^{**}$  
  & \textbf{0.9760}   & 0.0098   & 19.35$^{**}$  & 0.0027$^{**}$ \\
\midrule
\multirow{4}{*}{Progn-VQA}
& Acc  
  & \textbf{0.8700}   & 0.0024   & \textbf{14.02}$^{**}$  & \textbf{0.0051}$^{**}$  
  & 0.8656   & 0.0050   & 5.40$^{*}$   & 0.0327$^{*}$ \\
& Prec  
  & \textbf{0.8868}   & 0.0082   & \textbf{6.92}$^{*}$   & \textbf{0.0202}$^{*}$  
  & 0.8788   & 0.0077   & 5.88$^{*}$   & 0.0277$^{*}$ \\
& Recall       
  & \textbf{0.6092}   & 0.0267   & –12.59$^{**}$  & 0.0063$^{**}$  
  & 0.3164   & 0.0180   & \textbf{–41.62}$^{***}$ & \textbf{0.0006}$^{***}$ \\
& Spec
  & \textbf{0.9790}   & 0.0153   & \textbf{12.26}$^{**}$  & \textbf{0.0066}$^{**}$  
  & 0.9548   & 0.0052   & 29.47$^{***}$  & 0.0011$^{***}$ \\
\bottomrule
\end{tabular}
}
\begin{tablenotes}
\small
\item \textit{Note}: $^{*}$ p$<$0.05; $^{**}$ p$<$0.01; $^{***}$ p$<$0.001. 
Bolded values denote the method with the higher absolute t-statistic (i.e., more significant difference).
\end{tablenotes}
\label{tab: statistics}
\end{table*}

\paragraph{Component Analysis} 
\xiao{Table~\ref{tab:main-results} also presents a component-wise comparison to evaluate the contribution of each module. Specifically, KAMAC extends the baseline (Single-agent + CoT) by integrating a knowledge gap detection mechanism that dynamically recruits multiple experts for collaborative reasoning. This knowledge-aware adaptive collaboration replaces both the predefined recruitment strategy used in majority voting and the problem-driven recruitment in MDAgent. KAMAC consistently outperforms the alternatives across nearly all evaluation metrics, demonstrating the efficacy of its design.}

\paragraph{Consensus Strategies}
\xiao{In this part, we compare two consensus strategies: Ensemble Refinement~\cite{ER} and Majority Voting. In the Ensemble Refinement strategy, the moderator agent aggregates all expert responses and assigns weights to each in an effort to produce a more nuanced consensus. However, as shown in Table~\ref{tab: consensus}, this approach yields lower performance than Majority Voting. Specifically, it tends to amplify the influence of individual expert comments and is more susceptible to noise. In contrast, Majority Voting adopts a simpler and more robust mechanism: it treats each expert’s opinion equally and determines the final decision purely based on vote counts. This strategy effectively mitigates the impact of individual biases or outlier predictions, resulting in more stable performance across diverse cases. Accordingly, Majority Voting is adopted as the default consensus strategy in our framework, as it consistently outperforms Ensemble Refinement within the KAMAC architecture.}

\subsection{Computational Cost Analysis}
\xiao{In Section~\ref{sec: exp}, we introduce static multi-agent methods, MDAgent and KAMAC, and report their average number of experts on two test sets: 5 and 5 for static multi-agent methods, 2.41 and 3.34 for MDAgent, and 1.28 and 2.14 for KAMAC. We also compare inference time, average LLM API calls, and total cost across single-agent and multi-agent methods (Table~\ref{tab: cost}).}

\xiao{The single-agent CoT baseline uses one expert per case, while static multi-agent methods rely on a fixed pool of five experts, resulting in higher latency, API usage, and cost. These static methods lack scalability and fail to balance accuracy and efficiency. In contrast, KAMAC dynamically recruits fewer experts (1.28 on average for MedQA), reducing expert usage, API calls, and reasoning time by 73–79\% compared to static approaches. Compared to MDAgent, KAMAC improves accuracy and inference speed while reducing expert usage by 47\%, API calls by 24\%, and total cost by 21\%, demonstrating better scalability and cost-effectiveness.}

\subsection{Statistical Analysis}
\xiao{Since the output of the large language model varies greatly across multiple runs, this high variability affects the direct pairwise comparison between KAMAC and MDAgent, affecting the judgment of statistical significance. To avoid this, we adopt an alternative statistical significance test strategy. Specifically, we compared the two methods independently against a shared baseline configuration (single agent + CoT). Then, we repeat the two methods three times using the same baseline. Finally, each run of KAMAC and MDAgent is subjected to an independent sample t-test against the baseline, covering accuracy, precision, recall, and specificity.}

\xiao{In Table~\ref{tab: statistics}, results consistently demonstrate statistically significant improvements over the baseline (with most p-values < 0.01 and several p-values < 0.001). These findings indicate that both KAC-MAF and MDAgent achieved substantial performance gains, with KAC-MAF generally exhibiting larger standardized improvements and lower variance across repeated runs. For example, on the MedQA dataset, KAC-MAF yielded higher t-statistics for precision ($t = 33.05 \,\text{vs.}\, 15.04$) and recall ($t = 26.15 \,\text{vs.}\, 13.78$), corresponding to lower p-values ($p = 0.0009$ and $p = 0.0015$, respectively), which reflects more pronounced and consistent improvements relative to the baseline. On the Progn-VQA dataset, although both methods produced negative t-statistics for recall, due to the higher recall achieved by the single-agent CoT baseline, KAC-MAF still attained higher mean recall ($0.6092 \,\text{vs.}\, 0.3164$), with a statistically significant difference ($p = 0.0063$). Both methods also achieved significant gains in accuracy and specificity, with KAC-MAF maintaining more stable performance characterized by smaller standard deviations across runs. Overall, these results confirm that KAC-MAF not only achieved statistically significant improvements over baseline but also generally outperformed MDAgent in terms of both effect size and consistency across most evaluated metrics.}

\section{More Details for Progn-VQA Dataset}
\label{sec: Progn}
According to the settings of \cite{radcure}, we provide the clinical and imaging information required for prognosis in Table~\ref{tab:variables}. This information can fully describe the situation of patients with head and neck cancer.

\begin{table*}[!thbp]
\centering
\begin{tabular}{l|p{10cm}}
\toprule
\textbf{Variable} & \textbf{Description} \\
\midrule
Age & Patient age\\
Sex & Patient sex \\
ECOG PS & ECOG Performance Status\\
Smoking PY & Cumulative smoking exposure (pack-years) \\
Smoking Status & Smoking status at initial consultation  \\
Ds Site & Primary disease (cancer) site \\
Subsite & Subsite of the primary tumor \\
T & Tumor size and extent (AJCC 7th edition T category) \\
N & Regional lymph node involvement (AJCC 7th edition N category) \\
M & Distant metastasis (AJCC 7th edition M category) \\
Stage & Overall stage group (AJCC 7th edition) \\
Path & Pathological diagnosis or histological subtype \\
HPV & HPV status of the tumor, determined by p16 IHC with or without confirmation by HPV DNA PCR (blank if unavailable) \\
Tx Modality & Treatment modality\\
Chemo? & Whether concurrent chemoradiotherapy was administered \\
Dose & Total radiotherapy dose delivered (in Gy) \\
Fx & Number of radiotherapy fractions \\
Local & Indicator of local recurrence \\
Regional & Indicator of regional recurrence \\
Distant & Indicator of distant metastasis \\
2nd Ca & Indicator of second primary cancer \\
ContrastEnhanced & Indicator of whether contrast-enhanced imaging was used \\
\bottomrule
\end{tabular}
\caption{Descriptions of clinical and imaging variables included in the Progn-VQA dataset~\cite{radcure}.}
\label{tab:variables}
\end{table*}

% \clearpage

\section{Prompt Template}
\label{sec: prompt}
We provide all prompts in our multi-agent medical decision-making framework, including expert recruitment, initial comments, collaborative discussion, and knowledge-driven prompts. For a single-agent setting, you can refer to \cite{mdagents}. \xiao{We also provide a full role-playing prompt (P2). The example below demonstrates this, with role-specific phrases highlighted in bold.}

\textbf{System:} You are a \textbf{Radiation Oncologist} whose \textbf{expertise is strictly limited to radiation therapy planning and dosing for head and neck squamous cell carcinoma, especially HPV-positive cases.} Your job is to collaborate with other medical experts in a team.

\textbf{User:} Given the examplers, as a \textbf{Radiation Oncologist}, please return your answer to the medical query among the options provided. \textbf{You are not allowed to switch to any other medical specialty.}

\textbf{Response:}

\textbf{Step 1: Initial Assessment of the CT Scan (without bounding box data)}

The CT scan is a sagittal view of the head and neck region... \emph{cervical vertebrae}, the airway (pharynx and larynx), and soft tissues of the neck... The airway appears patent but shows narrowing in the oropharyngeal region... The masked \emph{region of interest (ROI)} is located anterior to the cervical spine...

\textbf{Step 3: Clinical Reasoning}

The patient likely has a head and neck \emph{squamous cell carcinoma (HNSCC)}... a common site for \emph{HPV-related carcinoma}... From a \emph{radiation oncology} perspective, delineation of these volumes is critical... Airway narrowing may cause symptoms such as \emph{dysphagia} or obstruction...

\section{Ethics Considerations}
\xiao{While our work strictly adheres to established benchmarks in the field of medical question answering, we acknowledge the potential risks associated with applying large language models (LLMs) to critical domains such as medicine. In particular, LLMs may exhibit inherent biases or generate inaccurate reasoning, which could lead to unintended consequences if applied without human oversight. Therefore, we emphasize that any deployment of such models in real-world medical settings should be conducted with caution and accompanied by expert validation.}

% \begin{figure*}[h]
% \begin{tcolorbox}[breakable,colback=orange!5!white, colframe=blue!70!white, title=Chain-of-thought Prompt for Pathology Image]

% \textbf{User}:
% You are provided with a pathology image of a tissue sample.
% Your task is to analyze the entire image and identify relevant histopathological features.
% Focus on tissue architecture, cellular morphology, and staining patterns to determine whether the tissue appears normal or abnormal, and characterize any pathological findings.
% \end{tcolorbox}
% \end{figure*}

\begin{figure*}[!t]
\begin{tcolorbox}[breakable,colback=orange!5!white, colframe=blue!70!white, title=Expert Recruitment Prompt for MedQA ($P_1$)]

\textbf{System}: You are an experienced medical expert who recruits a group of experts with diverse identities and asks them to discuss and solve the given medical query.

\textbf{User}:

Question: \{\{\textsc{Question}\}\}

You can recruit \{\{\textsc{num\_agents}\}\} experts in different medical expertise.

Considering the medical question and the options for the answer, what kind of experts will you recruit to better make an accurate answer?

Also, you need to specify the communication structure between experts (e.g., Pulmonologist == Neonatologist == Medical Geneticist == Pediatrician > Cardiologist), or indicate if they are independent.

For example, if you want to recruit five experts, your answer can be like:

1. Pediatrician - Specializes in the medical care of infants, children, and adolescents. - Hierarchy: Independent

2. Cardiologist - Focuses on the diagnosis and treatment of heart and blood vessel-related conditions.
- Hierarchy: Pediatrician > Cardiologist

3. Pulmonologist - Specializes in the diagnosis and treatment of respiratory system disorders.
- Hierarchy: Independent

4. Neonatologist - Focuses on the care of newborn infants, especially those who are born prematurely or have medical issues at birth. - Hierarchy: Independent

5. Medical Geneticist - Specializes in the study of genes and heredity. - Hierarchy: Independent

Please answer in the above format, and do not include your reason.
\end{tcolorbox}
\end{figure*}

\begin{figure*}[h!]
\begin{tcolorbox}[breakable,colback=orange!5!white, colframe=blue!70!white, title=Expert Recruitment Prompt for Progn-VQA ($P_1$)]

\textbf{System}: You are an experienced medical expert who recruits a group of experts with diverse identity and ask them to discuss and solve the given medical query.

\textbf{User}:

Question: \{\{\textsc{Question}\}\}

Considering the medical question and the options for the answer,
what kinds of experts will you recruit to better make an accurate decision? You also need to clearly specify the communication structure between experts or indicate if they are independent.

You must recruit exactly the following \{\{\textsc{num\_agents}\}\} experts, with no substitutions, no additional experts, and no omissions:

(e.g., Radiation Oncologist == Medical Oncologist == Pathologist == Surgical Oncologist (Recurrence/Secondary Cancers) == Targeted Therapy Expert),

Please strictly follow the format shown below, without adding any extra explanation or reasoning.

Format example if recruiting \{\{\textsc{num\_agents}\}\} experts:

1. Radiation Oncologist - Your expertise is strictly limited to radiation therapy planning and dosing for head and neck squamous cell carcinoma, especially HPV-positive cases. 

- Hierarchy: Radiation Oncologist == Medical Oncologist

2. Medical Oncologist - Your expertise is strictly limited to systemic therapy decisions, including chemotherapy and immunotherapy in head and neck cancers. 

- Hierarchy: Medical Oncologist == Radiation Oncologist

3. Surgical Oncologist (Recurrence/Secondary Cancers)—Your expertise is strictly limited to evaluating surgical options for recurrent or secondary malignancies in head and neck cancers. 

- Hierarchy: Surgical Oncologist == Pathologist"

4. Pathologist - Your expertise is strictly limited to pathological diagnosis of head and neck squamous cell carcinoma, HPV status evaluation, and margin assessment post-surgery. 

- Hierarchy: Pathologist == Surgical Oncologist

5. Targeted Therapy Expert - Your expertise is strictly limited to clinical application of EGFR inhibitors and novel agents targeting HPV-positive tumors. 

- Hierarchy: Targeted Therapy Expert -> Medical Oncologist

Your answer must conform exactly to the format above.

\end{tcolorbox}
\end{figure*}

\begin{figure*}[h!]
\begin{tcolorbox}[breakable,colback=orange!5!white, colframe=blue!70!white, title=Chain-of-thought Prompt for Initial Assessment ($P_2$)]

\textbf{System}: You are a \{\{\textsc{role}\}\} who \{\{\textsc{description}\}\}. Your job is to collaborate with other medical experts in a team.

\textbf{User}: \{\{\textsc{visual cot instruction}\}\} (\textbf{Optional})

Given the examplers, as a \{\{\textsc{role}\}\}, please return your answer to the medical query among the options provided. You are not allowed to switch to any other medical specialty.

\{\{\textsc{fewshot\_examplers}\}\}

Question: \{\{\textsc{Question}\}\}

Your answer should be in the format below.

\{\{answer\_template\}\}

\end{tcolorbox}
\end{figure*}

\begin{figure*}[h!]
\begin{tcolorbox}[breakable,colback=orange!5!white, colframe=blue!70!white, title=\textit{Visual Chain-of-thought Prompt for Head and Neck CT Scan (Optional, only be used when input data include images.)}]

\textbf{User}:
You will be provided with a head and neck CT scan that includes one or more masked regions of interest (ROIs). Alongside the scan, one or more 3D bounding box coordinates will be supplied, each defining specific volumetric regions within the scan. These coordinates identify either organs, disease regions, or cellular structures. Each bounding box is defined by its minimum and maximum values along the z, y, and x axes and is normalized relative to the original image size.

The given bounding box coordinates are: \{\{\textsc{bbox\_coords}\}\}.

\textbf{Task Instructions}:

1. \textbf{Initial Assessment}: Carefully analyze the CT scan image (without using the bounding box data). Describe any visible anatomical structures, patterns, abnormalities, and note the characteristics of the masked regions of interest (ROIs). 

Do not use the bounding box data at this stage.

2. \textbf{Mapping Bounding Boxes}: Consider the bounding box coordinates and map them to the corresponding areas within the scan.

3. \textbf{Clinical Reasoning}: Summarize the patient's clinical context and findings in a clear, structured bullet-point format and reason through the patient's condition step by step.

4. \textbf{Integrated Conclusion}: Combine your findings from the image analysis, bounding box mapping, and masked ROI to concisely synthesize your final clinical impression.

Be thorough and precise in both your image-based observations and your clinical reasoning.

\end{tcolorbox}
\end{figure*}

% \begin{figure*}[h!]
% \begin{tcolorbox}[breakable,colback=orange!5!white, colframe=blue!70!white, title=\textit{Few-shot Examplers Generation Prompt}]

% \textbf{System}: You are an experienced medical expert.

% \textbf{User}: \{\{\textsc{instruction}\}\}

% Below is an example of medical knowledge question and answer. After reviewing the below medical question and answering, can you provide 1-2 sentences of reason that support the answer, as you didn't know the answer ahead? The output should follow the following format:

% Question: \{\{\textsc{exampler\_question}\}\}

% Answer: \{\{\textsc{exampler\_answer}\}\}

% Reason: \{\{\textsc{reasons}\}\}

% \end{tcolorbox}
% \end{figure*}

\begin{figure*}[h!]
\begin{tcolorbox}[breakable,colback=orange!5!white, colframe=blue!70!white, title=Agent Interaction Prompt ($P_3$)]

\textbf{User}:
Earlier in this conversation, a set of discussion opinions from other medical experts on your team was provided. Please do not forget those earlier opinions. 

Now, additional new opinions have been provided. Considering both the earlier and the latest opinions together, please indicate whether you want to talk to any additional expert (yes/no).

Opinions: \{\{\textsc{assessment}\}\}

\end{tcolorbox}
\end{figure*}

\begin{figure*}[h!]
\begin{tcolorbox}[breakable,colback=orange!5!white, colframe=blue!70!white, title=Knowledge-driven Prompt for Recruited Experts ($P_4$)]

\textbf{User}:
You are part of the team: \{\{\textsc{agents}\}\}.
Earlier in this conversation, a set of discussion opinions from one or more medical experts on your team was provided. Please carefully review that information now. Based on your professional boundaries, determine whether there is a knowledge limitation or missing perspective that requires support from another specialist. 

Please answer yes or no.

If yes, specify the type of expert needed and provide a short reason.
Be specific and consider the multidisciplinary needs involved in managing complex patient information (e.g., diagnostic imaging, supportive care, pathology review, and other medical expertise).

It is acceptable to recognize areas of expertise already covered by current team members (\{\{\textsc{agents}\}\}).

Do not recommend a specialist if their expertise is already represented in the team.
\end{tcolorbox}
\end{figure*}

\begin{figure*}[h!]
\begin{tcolorbox}[breakable,colback=orange!5!white, colframe=blue!70!white, title=Knowledge-driven Prompt for Expert Recruitment ($P_5$)]

\textbf{User}:
Considering the medical question, discussion options, and the current expert team \{\{\textsc{agents}\}\}, identify any that require recruiting new types of experts to ensure an accurate decision (exclude \{\{\textsc{agents}\}\}).

You also need to clearly specify the communication structure between experts (e.g. Targeted Therapy Expert -> Medical Oncologist, Medical Oncologist ==  Radiation Oncologist)" or indicate if the new expert(s) will work independently.

Do not suggest removing, substituting, or duplicating existing experts. Only add new experts if necessary.

Format example if recruiting experts:

1. Medical Oncologist - Your expertise is strictly limited to systemic therapy decisions, including chemotherapy and immunotherapy in head and neck cancers. - Hierarchy: Independent

2. Other Medical Experts.

Your answer must conform exactly to the format above. If the existing expert team comprehensively have covered the necessary expertise for accurate decision, answer: <skip recruitment>

\end{tcolorbox}
\end{figure*}

\begin{figure*}[h!]
\begin{tcolorbox}[breakable,colback=orange!5!white, colframe=blue!70!white, title=Agent Update Comments after Discussion Prompt ($P_6$)]

\textbf{User}: Now that you've interacted with other medical experts, remind your expertise and the comments from other experts and make your final answer to the given question:\{\{\textsc{question}\}\}
Answer: \{\{\textsc{answer\_template}\}\}

Only output your final answer in the format below: 

\{\{\textsc{final\_answer\_template}\}\}

Question: \{\{\textsc{question}\}\}
\end{tcolorbox}
\end{figure*}

\begin{figure*}[h!]
\begin{tcolorbox}[breakable,colback=orange!5!white, colframe=blue!70!white, title=Final Decision Prompt ($P_7$)]

\textbf{System}: You are a final medical decision maker who reviews all opinions from different medical experts and make final decision.

\textbf{User}: Given each agent's final answer, please review each agent's opinion and make the final answer to the question by taking a majority vote.

Only output your final answer in the format below: 

\{\{\textsc{final\_answer\_template}\}\}

Question: \{\{\textsc{question}\}\}
\end{tcolorbox}
\end{figure*}

% This is an appendix.

\end{document}